\documentclass[lettersize,journal]{IEEEtran}
\usepackage{amsmath,amsfonts}
\usepackage{algorithmic}
\usepackage{algorithm}
\usepackage{array}
\usepackage[caption=false,font=normalsize,labelfont=sf,textfont=sf]{subfig}
\usepackage{textcomp}
\usepackage{stfloats}
\usepackage{url}
\usepackage{verbatim}
\usepackage{graphicx}
\usepackage{cite}
\usepackage[table]{xcolor} 
\usepackage{booktabs} 
\usepackage{multirow}
\usepackage{booktabs}
\usepackage{graphicx} 
\usepackage[colorlinks=true]{hyperref}

\hypersetup{
    colorlinks=true,
    citecolor=blue,     
    linkcolor=blue,     
}

\definecolor{firstcolor}{HTML}{FDAE6B}
\definecolor{secondcolor}{HTML}{FDD0A2}
\definecolor{thirdcolor}{HTML}{FEE6CE}
\definecolor{restcolor}{HTML}{FFF5EB}

\begin{document}

\title{Dehaze-GaussianImage: Zero-Shot Dehazing via\\Efficient 2D Gaussian Splatting Representation}

\author{Yuhan~Chen, 
        Wenxuan~Yu, 
        Guofa~Li,~\IEEEmembership{Senior Member,~IEEE}, 
        Kunyang~Huang, 
        Ying~Fang, 
        Yicui~Shi, 
        Wenbo~Chu, 
        and Keqiang~Li

\thanks{This work was supported by the National Natural Science Foundation of China under Grant No. 52272421. (\textit{Corresponding author: Wenbo Chu.})}%
\thanks{Yuhan~Chen, Wenxuan~Yu, Guofa~Li, Ying~Fang, and Yicui~Shi are with the College of Mechanical and Vehicle Engineering, Chongqing University, Chongqing 400044, China (e-mail: 20240701028@stu.cqu.edu.cn; wenxuanyu@cqu.edu.cn; liguofa@cqu.edu.cn; yingfang@stu.cqu.edu.cn; yicuishi@cqu.edu.cn).}%
\thanks{Kunyang Huang is with the Department of Electrical and Computer Engineering, Carnegie Mellon University, Moffett Field, CA 94035, USA (e-mail: kunyangh@andrew.cmu.edu).}%
\thanks{Wenbo~Chu is with the National Innovation Center of Intelligent and Connected Vehicles, Beijing 100089, China (e-mail: chuwenbo@wicv.cn).}%
\thanks{Keqiang~Li is with the School of Vehicle and Mobility, Tsinghua University, Beijing 100084, China (e-mail: likq@tsinghua.edu.cn).}}

\markboth{IEEE TRANSACTIONS ON COMPUTATIONAL IMAGING}%
{Shell \MakeLowercase{\textit{et al.}}: A Sample Article Using IEEEtran.cls for IEEE Journals}


\maketitle

\begin{abstract}
Existing single image dehazing methods are often constrained by computational redundancy in pixel-level optimization and the lack of physical interpretability in implicit neural networks. These limitations hinder the balance between representation efficiency and reconstruction fidelity. To address these issues, we propose Dehaze-GaussianImage, the first zero-shot framework that introduces 2D Gaussian Splatting (2DGS) into the image dehazing domain to break the traditional pixel-grid processing paradigm. Distinct from static convolutional neural networks (CNNs) or Transformers, our approach models hazy images as continuous and dynamically evolvable anisotropic Gaussian fields. Specifically, we propose a novel reconstruction-decoupling zero-shot learning strategy that embeds the atmospheric scattering model into the Gaussian parameter space. This strategy drives Gaussian primitives to adaptively split, clone, and prune during optimization, achieving geometric-level decoupling of the transmission medium and clear textures. Furthermore, explicit structure-preserving constraints are introduced to suppress artifacts commonly caused by traditional physical priors. Experimental results demonstrate that the proposed method achieves state-of-the-art (SOTA) performance in a fully unsupervised manner with minimal parameters, highlighting the potential of explicit Gaussian representation for low-level vision tasks. 
\end{abstract}


\section{Introduction}
\IEEEPARstart{H}{igh-quality} image input is essential for the robustness of downstream tasks in computer vision systems, such as object detection, semantic segmentation, and autonomous driving. However, adverse weather conditions, including fog and haze, often lead to severe contrast degradation, color shifts, and loss of structural details in captured images. This degradation process is typically formulated by the atmospheric scattering model:
\begin{equation}
I(x) = J(x)t(x) + A(1 - t(x))
\label{eq:scattering_model}
\end{equation}
where $I(x)$ is the observed hazy image, $J(x)$ represents the latent clear scene radiance, $A$ denotes the global atmospheric light, and $t(x) = e^{-\beta d(x)}$ is the transmission map describing the ability of light to penetrate the medium. Single image dehazing aims to restore $J(x)$ from $I(x)$, which remains a highly ill-posed inverse problem since both $t(x)$ and $A$ are unknown~\cite{ref1}.

To address this challenge, early studies primarily relied on hand-crafted priors. The Dark Channel Prior (DCP) proposed by He et al.~\cite{ref12} represents a milestone in single-image dehazing. This method leverages the statistical regularity where at least one color channel in local patches of haze-free natural images possesses near-zero intensity. Based on this observation, computable transmission estimates are constructed. Although DCP performs effectively in texture-rich scenes, failure often occurs in regions lacking dark pixels such as the sky or highlights. These limitations result in over-dehazing and blocking artifacts. Subsequently, more robust statistical models like the Color Attenuation Prior (CAP)~\cite{ref50} were introduced to characterize the correlation between brightness, saturation, and scene depth. This extension addresses the deficiencies of DCP in bright areas and expands physical priors from dark pixel assumptions to color attenuation regularities. Nevertheless, prior-based methods remain sensitive to specific scene assumptions and frequently incur halo artifacts or color distortions~\cite{ref3}.

With the rapid development of deep learning, CNN-based methods have become dominant. DehazeNet~\cite{ref8} and AOD-Net~\cite{ref9} were among the first to estimate parameters by learning the mapping from hazy images to transmission maps. Subsequent architectures shifted toward end-to-end dehazing paradigms. For instance, FFA-Net~\cite{ref5} achieves SOTA performance through a feature fusion attention mechanism, while GridDehazeNet~\cite{ref11} addresses bottleneck issues via a multi-scale grid structure. Although supervised methods demonstrate superior performance on synthetic datasets, severe domain shifts are frequently encountered in real-world scenarios. This is primarily due to the difficulty of obtaining paired hazy and clear images~\cite{ref7,ref10}.

To alleviate the dependency on paired data, unsupervised and zero-shot dehazing methods have emerged as promising alternatives. For instance, CycleGAN~\cite{ref13} and the variant Cycle-Dehaze~\cite{ref14} utilize adversarial training to achieve domain adaptation. Contrastive learning is also introduced in UCL-Dehaze~\cite{ref21} to maximize the mutual information between clear images and dehazing results. Recently, impressive texture restoration has been achieved by diffusion-based approaches~\cite{ref22,ref23,ref24} through generative priors. However, most existing dehazing algorithms including CNNs and Transformers are fundamentally limited by discrete pixel grid representations. In high-resolution scenarios, significant computational redundancy is incurred by pixel-wise convolution or attention mechanisms. This pixel-level processing often neglects the continuous structural nature of image content. Furthermore, the lack of physical interpretability in neural network mappings makes it difficult to explicitly impose structure-preserving constraints during optimization.

Recently, 3D Gaussian Splatting (3DGS)~\cite{ref34} has revolutionized the field of novel view synthesis owing to its exceptional rendering speed and explicit scene representation. Inspired by this advancement, the dimensionality reduction of Gaussian splatting has been investigated for 2D image representation. Specifically, GaussianImage~\cite{ref30} and Large Images Are Gaussians~\cite{ref28} pioneered the use of anisotropic 2D Gaussian primitives to represent high-resolution images efficiently. Superior compression ratios and reconstruction quality were achieved through these frameworks. Subsequently, the potential of explicit representations for low-level vision tasks was further demonstrated by 2DGS-based methods in low-light enhancement~\cite{ref26,ref27} and super-resolution~\cite{ref33}. Nevertheless, current 2DGS applications are primarily restricted to image fitting or color adjustment. The application of 2DGS geometric properties to solve single image dehazing involving complex physical degradation models remains an unexplored area. Existing Gaussian-based dehazing approaches~\cite{ref46,ref47,ref49} focus on multi-view scene reconstruction and rely on 3D geometric consistency. These methods are therefore inapplicable to the widespread demand for single image processing.

\begin{figure}[!t]
\centering
\includegraphics[width=\columnwidth]{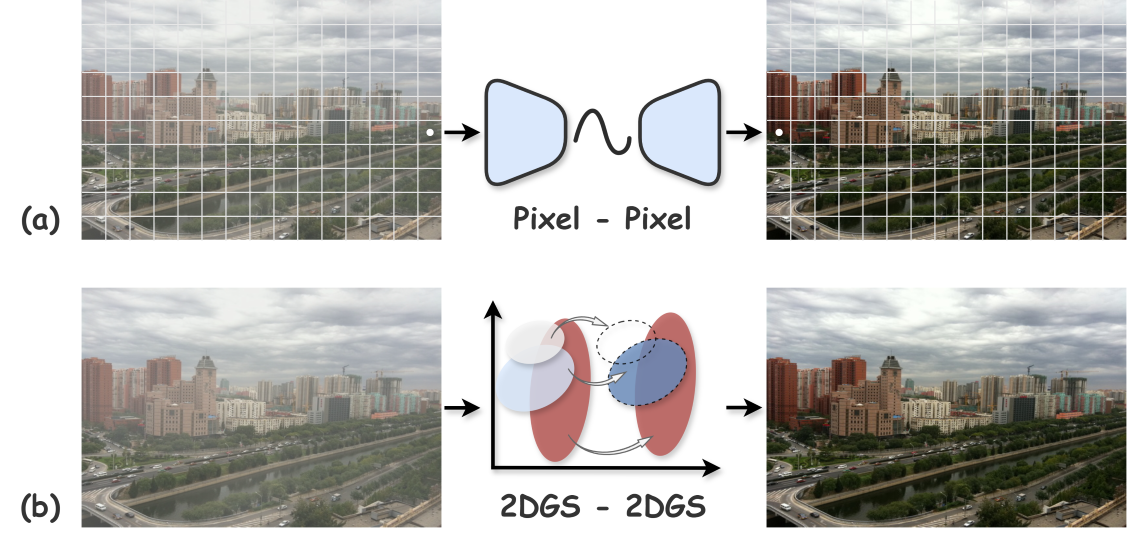}
\caption{Comparison of image dehazing paradigms across different representation domains. (a) Conventional pixel-domain dehazing schemes. (b) Proposed dehazing scheme in the 2DGS-based compressed representation domain. Dehaze-GaussianImage is the first framework to perform image dehazing directly within the 2DGS compressed representation space.}
\label{fig_1}
\end{figure}

To address these limitations, Dehaze-GaussianImage is proposed as the first zero-shot single-image dehazing framework based on 2DGS compressed representation. The dehazing process is reformulated as a physical-layer decoupling task within the Gaussian parameter space as illustrated in Fig.~\ref{fig_1}. This approach differs from traditional pixel-grid operations and is based on the insight that haze formation results from the interaction between scene structures and the transmission medium. Such coupling can be explicitly separated through the adaptive evolution of Gaussian primitives. By embedding the physics of atmospheric scattering into Gaussian attributes, the proposed framework enables the simultaneous reconstruction of clear textures and the estimation of the transmission medium in a continuous domain.

Specifically, the proposed framework consists of two key stages. In the first stage, the hazy image is encoded into a continuous 2D Gaussian field. Anisotropic covariance matrices are utilized in this process to accurately capture textures and edges. In the second stage, a reconstruction-decoupling optimization strategy is implemented. By explicitly embedding the atmospheric scattering model into the Gaussian rendering pipeline, a lightweight U-Net is designed to predict the transmission field. Meanwhile, Gaussian primitives are responsible for representing the latent clear scene. To address the ill-posed nature of unsupervised training, an adaptive density control mechanism based on physical features is introduced. This mechanism allows Gaussian primitives to perform splitting, cloning, and pruning automatically according to physical residuals during optimization. Consequently, the model adaptively focuses on high-frequency detail regions while smoothing dense haze areas. Furthermore, a set of physical loss functions including brightness and grayscale preservation is developed to resolve the failures of traditional physical models in sky regions and on white objects. The primary contributions are summarized as follows:

\begin{itemize}
\item We pioneer the integration of 2DGS into single image dehazing. By proposing a novel paradigm that transcends the inherent limitations of discrete pixel grids, our approach achieves a deep fusion between explicit image representation and physical degradation models.
\item We propose a novel zero-shot learning strategy that incorporates the atmospheric scattering model as a geometric constraint, driving Gaussian primitives to dynamically optimize their position, shape, and opacity, thereby achieving high-quality dehazing in the absence of paired data.
\item Experiments on multiple benchmark datasets demonstrate that Dehaze-GaussianImage surpasses existing unsupervised and even certain supervised methods in visual quality while maintaining minimal parameters and superior interpretability, which highlights the significant potential of explicit Gaussian representation in low-level vision tasks.
\end{itemize}

\section{Related Work}
Dehaze-GaussianImage is situated at the intersection of explicit neural representation and low-level vision tasks, specifically targeting the emerging paradigm of image dehazing within the compressed representation domain. To provide a comprehensive context, the related literature is categorized into three primary sections. First, we review the evolution of single image dehazing methods. Second, we discuss the advancements in 3DGS technology and its applications in inverse imaging. Finally, we highlight the recent progress in 2DGS-based image representation and processing.
\subsection{Single Image Dehazing}
Single image dehazing is a classic inverse problem in low-level vision that aims to recover clear scene radiance from degraded observations~\cite{ref1,ref3}. This task has long remained a focal point in the vision community. Based on the core paradigms, existing dehazing methods are generally categorized into traditional methods based on physical priors, deep network methods based on supervised learning, and unsupervised learning methods.

\textbf{Physical prior-based methods.}
Early dehazing efforts primarily rely on the observation of statistical characteristics in natural scenes. DCP~\cite{ref12} is recognized as a pioneering work in this field. This method is founded on the assumption that haze-free images possess at least one color channel with extremely low pixel values in non-sky regions, which facilitates the coarse estimation of transmission maps. Inspired by this, the Color Attenuation Prior (CAP)~\cite{ref50} was proposed to circumvent the limitations of DCP in bright regions by establishing a linear model between brightness, saturation, and depth of field. Although high interpretability is offered by these physics-based methods, tedious parameter tuning is often required. Furthermore, severe blocking artifacts and color distortions are frequently observed in scenes that violate the underlying prior assumptions, such as those containing large-scale sky areas or white objects.

\begin{figure*}[!t]
\centering
\includegraphics[width=\textwidth]{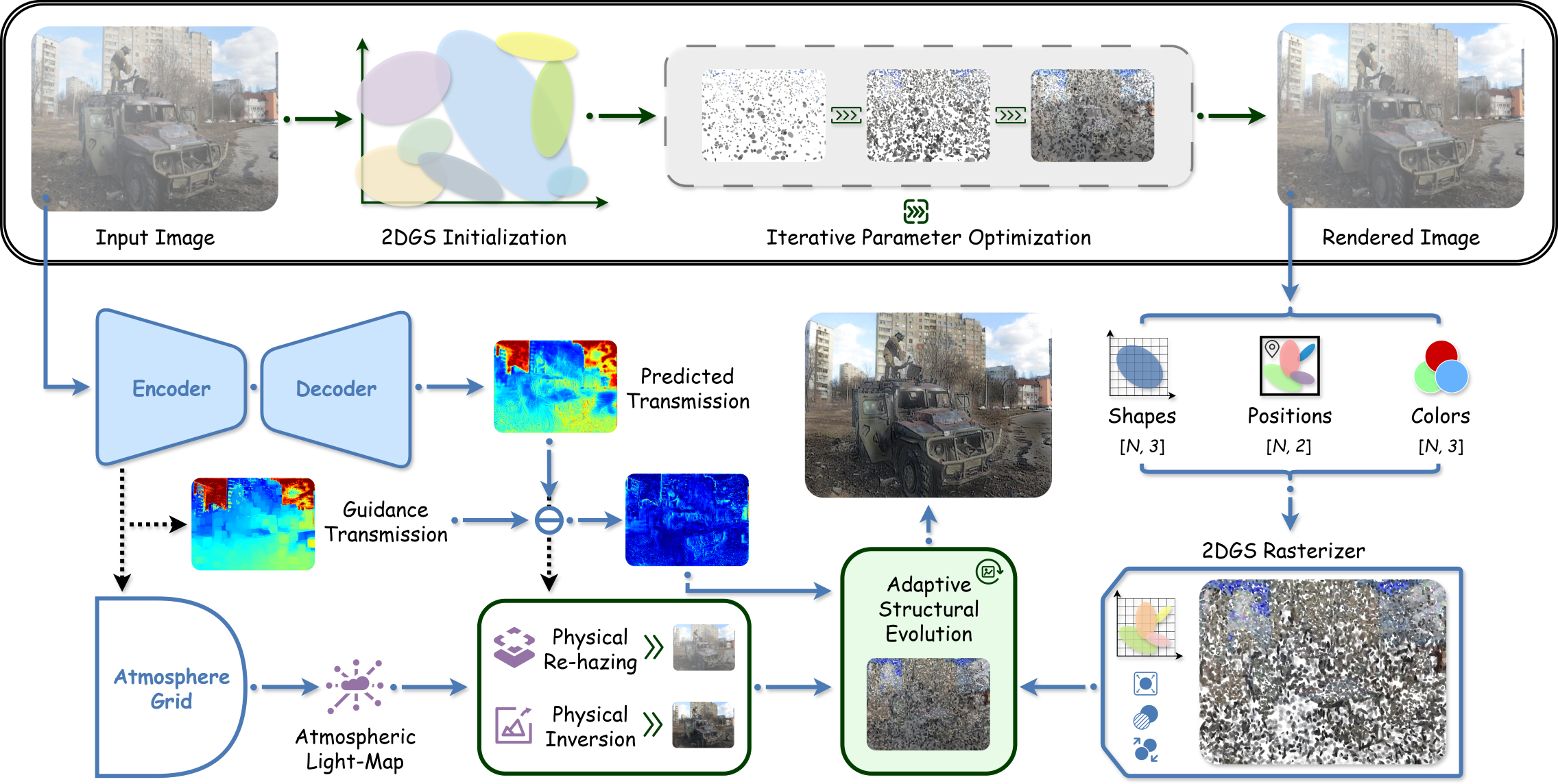}
\caption{Overview of the Dehaze-GaussianImage framework. The pipeline is composed of two parallel stages. First, the hazy image is reconstructed via self-supervised 2DGS to extract an explicit representation. Simultaneously, a joint optimization field is established based on the atmospheric scattering model (ASM). The hazy scene is decoupled into a 2D Gaussian field $J$ representing clean radiance, a predicted transmission map $t$, and an atmospheric light map $A$. Joint optimization is subsequently performed using reconstruction loss and physical prior constraints. High-fidelity zero-shot dehazing is achieved through this integrated process.}
\label{fig_2}
\end{figure*}

\textbf{Supervised learning-based methods.}
Driven by the development of CNN, data-driven approaches have gradually dominated the dehazing field. DehazeNet~\cite{ref8} and AOD-Net~\cite{ref9} represent early landmarks in this domain. Specifically, a CNN is employed by the former to learn the mapping from hazy images to transmission maps. In contrast, end-to-end dehazing is achieved by the latter through the reformulation of the atmospheric scattering model. To capture multi-scale features, a multi-scale CNN~\cite{ref10} was designed for coarse-to-fine transmission estimation. Following the evolution of network architectures, a feature fusion attention mechanism was introduced in FFA-Net~\cite{ref5} to adaptively process features across different channels and pixels. This mechanism significantly enhances restoration performance. Furthermore, bottleneck issues in multi-scale fusion were alleviated by GridDehazeNet~\cite{ref11} through a novel grid network that integrates pre-processing and post-processing modules. More recently, enhanced Pix2Pix networks were explored by Qu et al.~\cite{ref4}. HardGAN~\cite{ref6} was also developed to improve visual realism through adversarial training. To mitigate the domain shift between synthetic and real-world hazy images, physical priors are utilized by the PSD framework~\cite{ref7} to guide the adaptation from synthetic to real data. Additionally, a two-stage hybrid feature fusion network was proposed in TFFD-Net~\cite{ref2} to further refine detail restoration. Despite superior performance on synthetic datasets, these supervised methods remain heavily dependent on paired clear-hazy data. Generalization capabilities are often limited by the scarcity of such paired data in real-world scenarios.

\textbf{Unsupervised and zero-shot learning methods.}
To eliminate the dependency on paired data, unsupervised learning has emerged as a significant focal point. CycleGAN~\cite{ref13} and the optimized variant Cycle-Dehaze~\cite{ref14} leverage cycle-consistency loss to facilitate the translation from the hazy domain to the clear domain without paired supervision. However, hallucinated textures are frequently generated by GAN-based methods. To mitigate this issue, the decoupling of hazy images into content and haze features was explored in~\cite{ref15,ref18} to achieve dehazing through adversarial training. Contrastive learning has also been introduced to the field. Specifically, network optimization is driven by UCL-Dehaze~\cite{ref21} and AECR-Net~\cite{ref20} through the maximization of mutual information between clear images and restored results.

Recently, zero-shot learning has emerged as a prominent area of interest because it operates without pre-training. For instance, YoLY~\cite{ref16} utilizes a completely untrained neural network that is optimized directly on a single test image. Weak supervision and deep decomposition are further explored in RefineDNet~\cite{ref17} and self-augmentation methodologies~\cite{ref19} to enhance the robustness of unpaired dehazing. Zero-shot dehazing is even extended to surgical robotic vision in~\cite{ref25}. Driven by the advancement of generative models, diffusion-based priors are increasingly incorporated into dehazing frameworks. Specifically, pre-trained diffusion models are utilized as priors in~\cite{ref22} and~\cite{ref23} to guide real-world dehazing in the spatial and frequency domains, respectively. Schrödinger Bridge theory is further integrated in~\cite{ref24} to optimize dehazing trajectories. However, these methods are fundamentally restricted by discrete pixel-grid representations regardless of their underlying architectures. Significant computational redundancy is incurred by such representations in high-resolution scenarios. Furthermore, the explicit decoupling of scene structures and the transmission medium is hindered by the black-box nature of neural networks under the physical constraints of the atmospheric scattering model.

\begin{figure*}[!t]
\centering
\includegraphics[width=\textwidth]{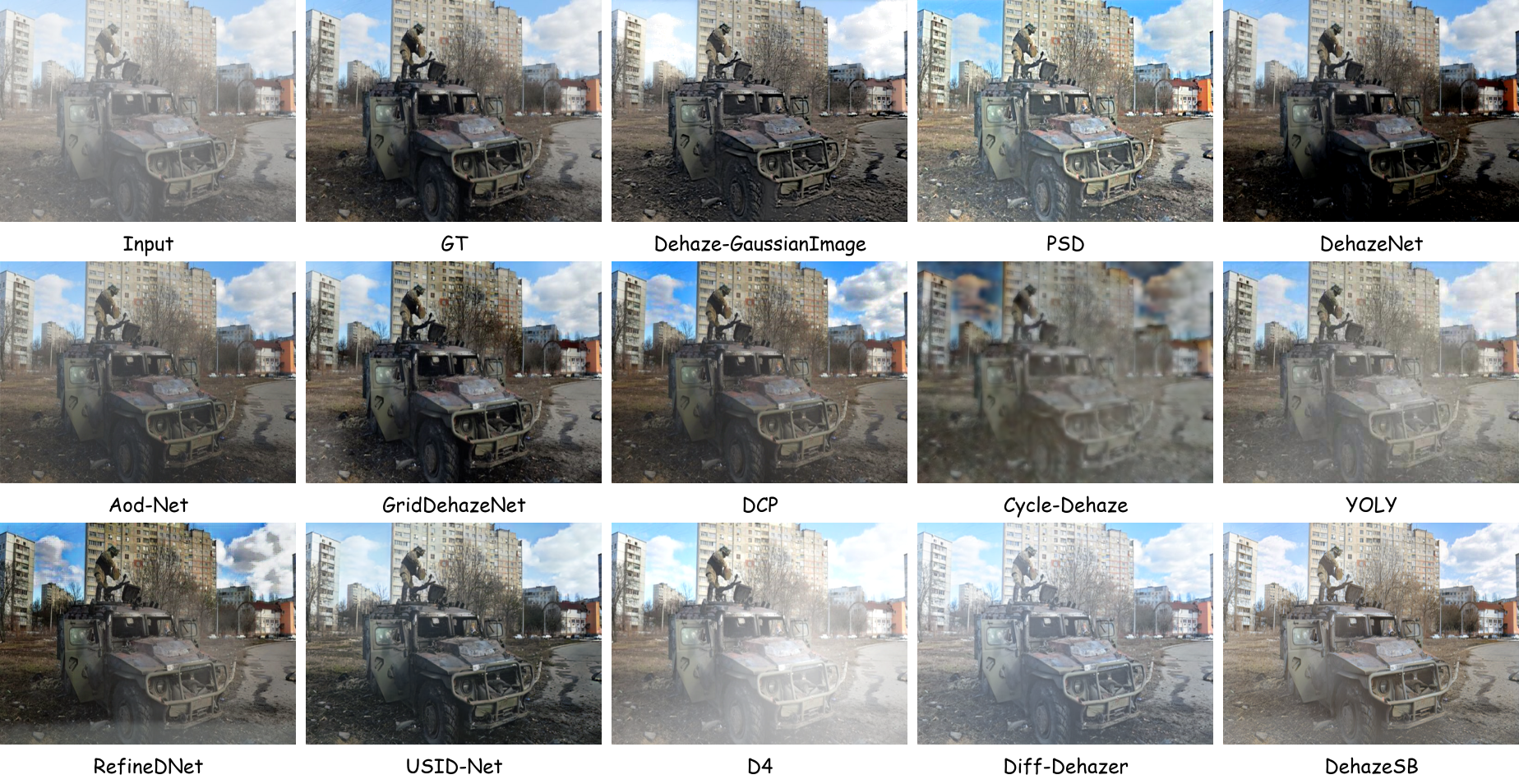}
\caption{Visual comparison between Dehaze-GaussianImage and SOTA methods on the NID dataset.}
\label{fig_3}
\end{figure*}

\subsection{Gaussian Splatting} 
3DGS~\cite{ref34} has emerged as a prominent explicit scene representation, significantly advancing the field of computer vision with real-time rendering speeds and high-fidelity reconstruction that surpass NeRF. Specifically, a scene is represented by a set of anisotropic 3D Gaussian primitives. These primitives are projected onto 2D image planes through a differentiable rasterization pipeline.

\textbf{Evolution and applications of Gaussian Splatting.} 
To address the geometric limitations of the original 3DGS, 2DGS~\cite{ref35} was introduced for more accurate surface reconstruction in radiance fields. 3DGS demonstrates significant potential in autonomous driving and large-scale scene modeling. Specifically, dynamic urban environments are modeled by Street Gaussians~\cite{ref36} and DrivingGaussian~\cite{ref40} through the effective capture of vehicle and pedestrian trajectories. In static large-scale scenarios, high-fidelity reconstruction is achieved by CityGaussian~\cite{ref42} and Momentum-GS~\cite{ref43} using tile-based parallel training and momentum self-distillation strategies. Furthermore, void issues in expansive scenes are mitigated by GaussianPro~\cite{ref41} via a progressive propagation mechanism. To accelerate training, sparse pixels and primitives are leveraged by Speedy-Splat~\cite{ref37} for ultra-fast convergence. Recent developments also integrate physical laws and generative models into Gaussian representations. For instance, physical constraints are incorporated into the splatting process by PMGS~\cite{ref38} and PEGS~\cite{ref39} to reconstruct complex motions over extensive spatio-temporal spans. In generative tasks, 2D and 3D diffusion models are bridged by GaussianDreamer~\cite{ref44} and ReconDreamer~\cite{ref45}, establishing a new paradigm for high-quality 3D scene generation from textual or visual prompts.

\textbf{Dehazing attempts based on Gaussian Splatting.} 
Inspired by the success of 3DGS in high-fidelity scene reconstruction, 3DGS has been adapted to process hazy environments. Specifically, haze components are decoupled during the optimization process by DehazeGS~\cite{ref46,ref47} and DehazeSplat~\cite{ref48} through the exploitation of multi-view consistency. This consistency is utilized to distinguish scene geometry from the suspended medium. Similarly, scene reconstruction in smoke-filled environments is addressed by SmokeSeer~\cite{ref49}. However, a significant limitation remains as these methods strictly require multi-view inputs to decouple haze and depth via motion parallax. Consequently, such geometry-based approaches are inapplicable to single image dehazing tasks where multi-view information is unavailable. A new physical dehazing paradigm is therefore necessitated to accommodate single image inputs by leveraging explicit 2D Gaussian representations.

\subsection{Gaussian Splatting-Based Image Representation.} 
Despite the success of 3DGS in novel view synthesis, its adaptation to 2D static image processing is challenging. Dimensional redundancy is exhibited by traditional 3DGS ellipsoids when representing planar images, and high-frequency textures are difficult to capture through direct projection. To address these issues, GaussianImage was introduced by Zhang et al.~\cite{ref30} as a framework for 2D image representation. In this method, an image is modeled as a continuous 2D Gaussian field. High-fidelity fitting is achieved by optimizing a set of anisotropic 2D Gaussian primitives defined by position, covariance, and color. Consequently, high-quality reconstruction is maintained with significantly fewer parameters than pixel grids, and a rendering speed of 1000 FPS is reached. This approach replaces the conventional pixel-based discrete representation paradigm, providing a fully differentiable and explicit alternative for image compression and low-level vision tasks.

\begin{figure*}[!t]
\centering
\includegraphics[width=\textwidth]{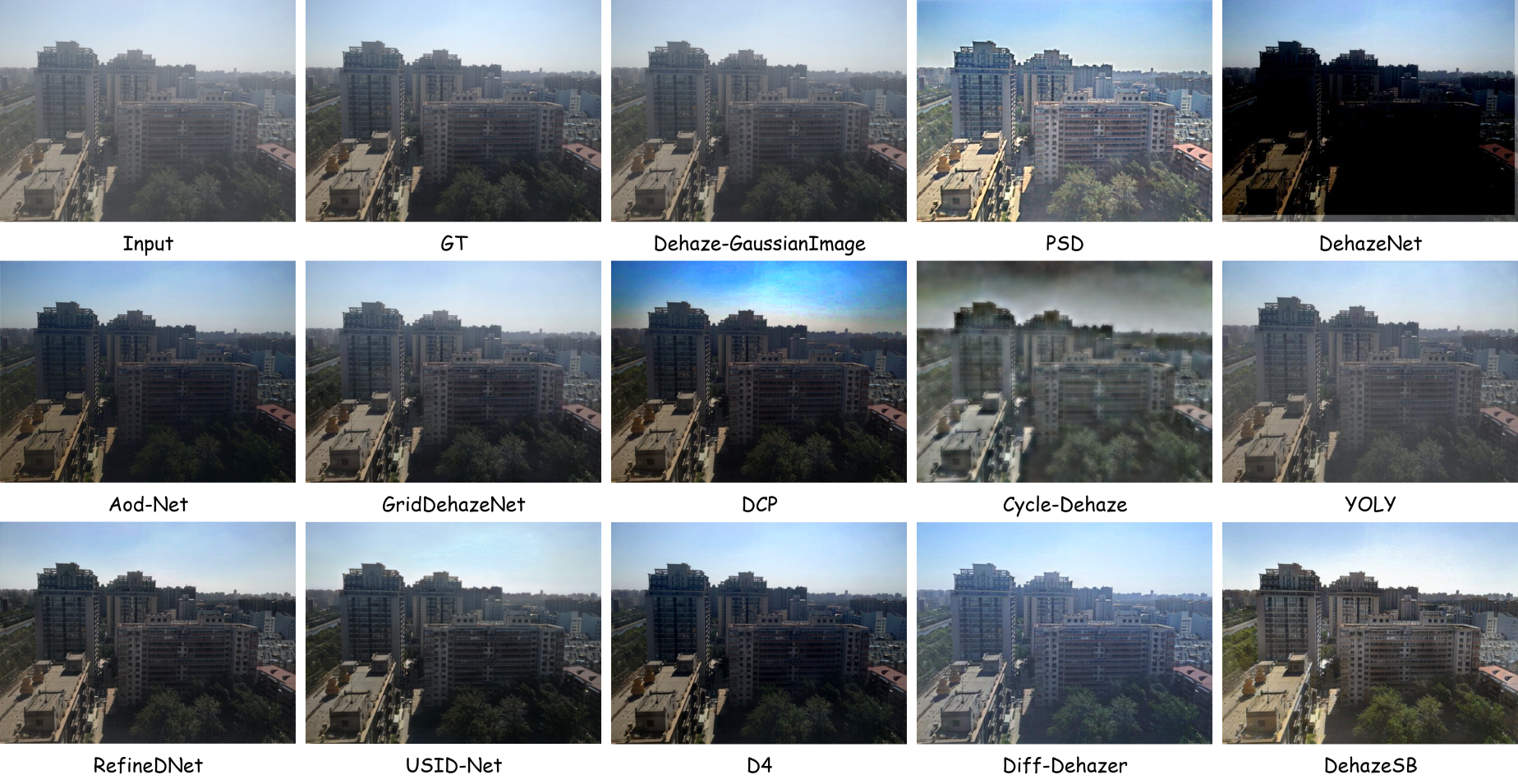}
\caption{Visual comparison between Dehaze-GaussianImage and SOTA methods on the SOTS dataset.}
\label{fig_4}
\end{figure*}

\textbf{Evolution and expansion of representation capabilities.} 
Driven by the inherent potential of 2DGS, subsequent efforts have focused on optimizing adaptability in complex scenarios. To address the representation challenges of ultra-high-resolution images, Zhu et al.~\cite{ref28} proposed Large Images Are Gaussians, which introduces a multi-level detail streaming strategy to facilitate fine-grained reconstruction under stringent memory constraints. Simultaneously, Instant GaussianImage~\cite{ref29} incorporates an adaptive quadtree mechanism, enabling Gaussian primitives to dynamically split and merge. This allows for the autonomous allocation of computational resources based on the structural complexity of the image content. The efficiency of this compressed representation even transcends conventional vision tasks. For instance, Jiang et al.~\cite{ref31} leveraged its sparsity for efficient dataset distillation, while Omri et al.~\cite{ref32} explored the cross-modal applications of compressed Gaussian latent spaces in vision-language alignment. These advancements collectively demonstrate the robust semantic capacity and versatility of explicit Gaussian representations.

\textbf{Applications and limitations in low-level vision tasks.} 
The transition from basic image fitting to complex low-level vision restoration is facilitated by the continuity and differentiability of 2DGS compressed representations. This shift from implicit grids to explicit primitives provides alternative strategies for the mitigation of specific degradation issues. Specifically, zero-shot brightness recovery is achieved by LL-GaussianImage~\cite{ref26} and LL-GaussianMap~\cite{ref27} through the adjustment of illumination gain attributes during optimization. Furthermore, the scale invariance of 2D Gaussians is utilized by GaussianSR~\cite{ref33} to perform image super-resolution at arbitrary scales. Artifacts typically associated with traditional CNN interpolation are consequently avoided.

While 2DGS demonstrates potential in global color adjustment and geometric resolution restoration, single image dehazing involves distinct physical challenges. Dehazing is characterized as a complex inverse problem where scene radiance and the transmission medium are highly coupled, extending beyond mere luminance enhancement or pixel interpolation. Existing 2DGS-based low-level vision methodologies lack explicit mechanisms for modeling the atmospheric scattering process. Consequently, haze density and object textures cannot be effectively distinguished during the optimization process.

To bridge this gap, we propose Dehaze-GaussianImage, the first framework that deeply embeds the atmospheric scattering model into the 2DGS compressed representation. As illustrated in Fig.~\ref{fig_2}, unlike previous methodologies, we introduce a reconstruction-decoupling strategy within the 2DGS domain. This strategy leverages physical constraints to drive the adaptive evolution of Gaussian primitives, effectively stripping the haze layer from the scene. Consequently, our approach achieves a seamless integration of explicit representation efficiency and physical dehazing mechanisms in a fully unsupervised manner.

\begin{figure*}[!t]
\centering
\includegraphics[width=\textwidth]{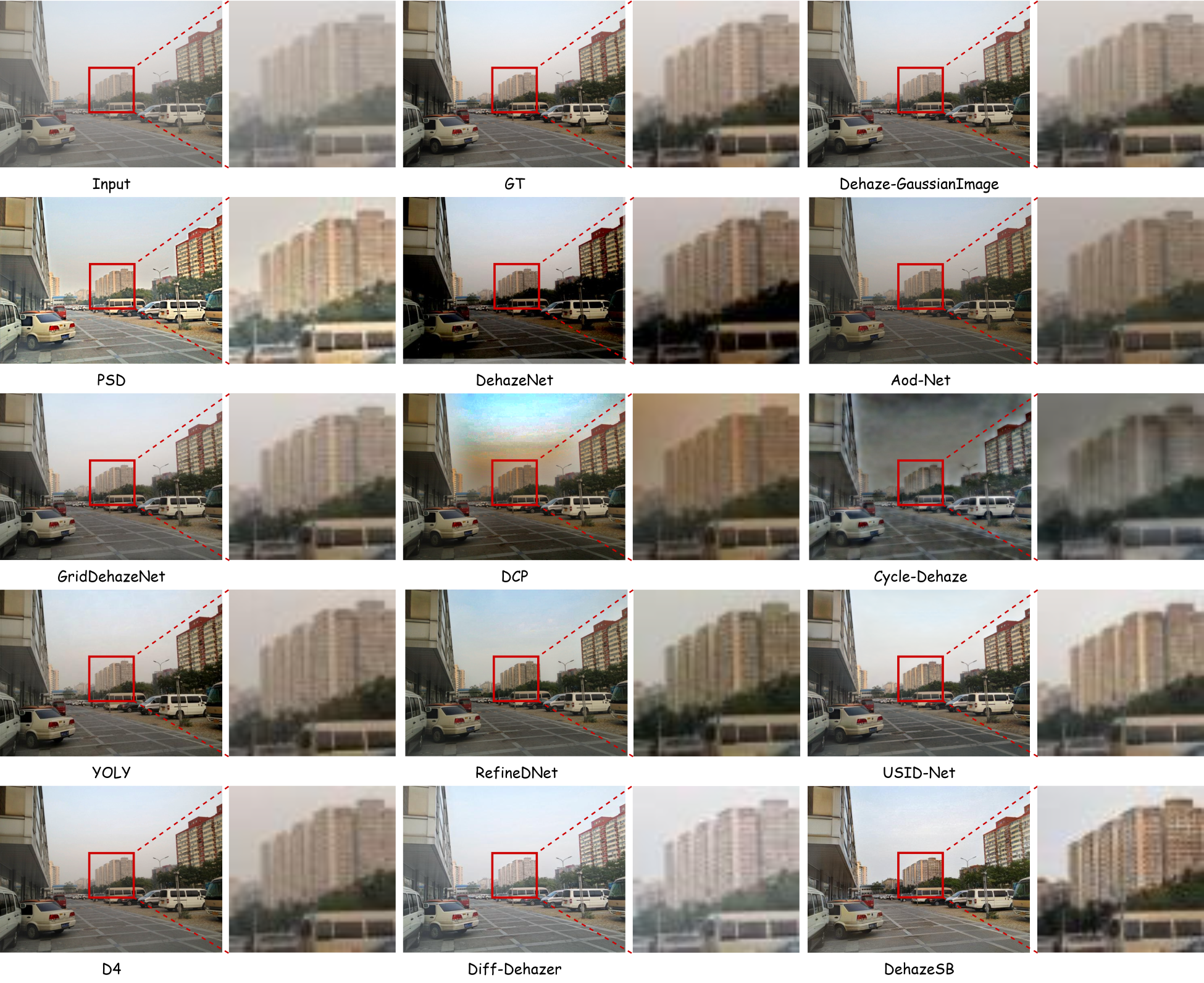}
\caption{Detailed feature comparison between Dehaze-GaussianImage and SOTA methods on the SOTS dataset. For each method, magnified views of the regions demarcated by red boxes are provided.}
\label{fig_5}
\end{figure*}

\section{Proposed Method}
Single image dehazing aims to restore the latent clear scene radiance $J$ from a degraded observation $I$ corrupted by atmospheric scattering. Following the classical atmospheric scattering model (ASM), the formation of a hazy image $I \in \mathbb{R}^{H \times W \times 3}$ is formulated as a linear convex combination of the clear scene $J$ and the global atmospheric light $A$:
\begin{equation}
I(x) = J(x)t(x) + A(1 - t(x)) \tag{2}
\end{equation}
where $x$ denotes the pixel coordinate and $t(x) \in [0, 1]$ represents the transmission map characterizing the medium's transmittance. This recovery process is inherently an ill-posed inverse problem, as it requires the simultaneous estimation of three unknown variables, namely $J$, $t$, and $A$, from only a single observation.

Existing dehazing approaches primarily leverage deep neural networks to perform end-to-end regression of $J$ or $t$ on discrete pixel grids. However, such grid-based representations suffer from two fundamental limitations. First, computational redundancy and low representation efficiency are observed, as dense convolutional operations must be performed over millions of pixels to restore high-frequency textures. This mechanism neglects the inherent sparsity of image content. Second, physical interpretability is limited because geometric structures and medium degradations are difficult to decouple within implicit feature spaces. Consequently, artifacts or color distortions are frequently introduced into the restored results.

To address these issues, the Dehaze-GaussianImage framework is proposed. As illustrated in Fig.~\ref{fig_2}, the traditional pixel-level prediction paradigm is replaced by 2DGS as the underlying scene representation. The proposed methodology comprises two cascaded optimization stages. First, in the Explicit Hazy Representation stage, 2D Gaussian primitives are utilized to encode the hazy image into an explicit representation characterized by high fidelity and a high compression ratio. Second, the Physics-Disentangled Evolution stage involves the construction of a physics-Gaussian decoupling loop. This loop facilitates the transition of the Gaussian field from a hazy state to a clear state. By integrating adaptive structural evolution with physical prior constraints, zero-shot dehazing is achieved without external supervision.

\subsection{Efficient Hazy Scene Representation via 2DGS}
In the first stage, the continuous hazy scene $I_{hazy}$ is parameterized into a discrete and differentiable set of anisotropic 2D Gaussian primitives $\mathcal{G}=\{g_1, g_2, \dots, g_N\}$.

\textbf{Gaussian Primitives and Parameterization.} 
Each 2D Gaussian primitive $g_i$ is characterized by a tuple of learnable parameters $\Theta_i = \{\mu_i, \Sigma_i, c_i, \alpha_i\}$ representing the center position, covariance matrix, RGB color vector, and opacity, respectively. On the image plane $\Omega$, the spatial response of the $i$-th Gaussian primitive at position $x$ is defined by the following probability density function:
\begin{equation}
G_i(x) = \exp\left(-\frac{1}{2}(x - \mu_i)^T \Sigma_i^{-1} (x - \mu_i)\right) \tag{3}
\end{equation}

To maintain the positive semi-definiteness of the covariance matrix $\Sigma_i$ and incorporate explicit geometric interpretations, $\Sigma_i$ is parameterized through a decomposition into rotation and scaling components. Specifically, the covariance matrix is formulated as:
\begin{equation}
\Sigma_i = R_i S_i S_i^T R_i^T \tag{4}
\end{equation}
where $R_i$ is constructed via the rotation angle $\theta_i$ and $S_i$ is defined by the axial scaling factors $s_{x}$ and $s_{y}$:
\begin{equation}
R_i = \begin{bmatrix} \cos \theta_i & -\sin \theta_i \\ \sin \theta_i & \cos \theta_i \end{bmatrix}, \quad 
S_i = \begin{bmatrix} s_{x,i} & 0 \\ 0 & s_{y,i} \end{bmatrix} \tag{5}
\end{equation}

This decomposition provides Gaussian primitives with the flexibility to stretch and rotate. Consequently, high-frequency structures including edges and textures are accurately captured. This anisotropic modeling capability surpasses the inherent limitations of conventional isotropic pixel grids.

\textbf{Splatting-based Differentiable Rasterization.} 
To project the Gaussian set onto the image plane, an efficient tile-based rasterizer is utilized. For a specific pixel $x$, the accumulated color $C(x)$ is formulated through a volume rendering process by integrating the contributions of Gaussians along the depth order:
\begin{equation}
C(x) = \sum_{i \in \mathcal{N}} c_i \alpha_i G_i(x) \prod_{j=1}^{i-1} (1 - \sigma_j G_j(x)) \tag{6}
\end{equation}
where $\mathcal{N}$ represents the sorted subset of Gaussians overlapping the pixel. The effective opacity $\sigma_i$ is obtained from the learnable opacity logits via a Sigmoid activation:
\begin{equation}
\sigma_i = \text{Sigmoid}(\sigma_i^{logits}) \tag{7}
\end{equation}

\textbf{Optimization and Compression Analysis.} 
During the first stage, the reconstruction error between the rendered image $I_{render}$ and the input hazy observation $I_{hazy}$ is minimized. To maintain pixel-level accuracy and perceptual structural consistency, the loss function is formulated as a weighted combination of the $\mathcal{L}_1$ norm and the Multi-Scale Structural Similarity (MS-SSIM) index:
\begin{equation}
\begin{split}
\mathcal{L}_{stage1} = & \lambda_{1} \|I_{render} - I_{hazy}\|_1 \\
& + \lambda_{ms\_ssim} \left(1 - \text{MS-SSIM}(I_{render}, I_{hazy})\right)
\end{split} \tag{8}
\end{equation}

\textbf{Compression Efficiency Analysis.} 
The 2DGS representation exhibits high data compression efficiency. For an image with a resolution of $H \times W$, the data complexity of a traditional pixel-based representation is $\mathcal{O}(H \cdot W)$. Significant storage savings are realized by constraining the number of Gaussians such that $N \ll H \cdot W$. For a high-definition image of $1920 \times 1080$, the parameter count is reduced to approximately 0.3\% of the conventional pixel matrix when $N$ is limited to 2000. This sparse representation reduces the computational load and provides inherent filtration of high-frequency perturbations caused by imaging noise. Consequently, a compact and structured geometric prior is established to facilitate more robust scene recovery in subsequent dehazing tasks.

\subsection{Physics-Disentangled Zero-Shot Framework}
The second stage constitutes the primary component of the Dehaze-GaussianImage framework. In this stage, the hazy Gaussian field obtained from the first stage is transformed into a decouplable generative model. A closed-loop system consisting of a Gaussian J-Estimator, a UNet T-Estimator, and a Physics Re-hazing Layer is proposed. This system is designed to disentangle the scene radiance $J$, the transmission map $t$, and the atmospheric light $A$ in a fully unsupervised manner.

\textbf{Tri-Branch Disentanglement Architecture.} 
To address the entanglement within the ASM, three independent and parallel parameter branches are incorporated into the Dehaze-GaussianImage framework.

\paragraph{the Scene Radiance Branch($\mathcal{G} \rightarrow J_{gs}$)} 
The modeling of the clear image $J$ is exclusively assigned to the Gaussian field $\mathcal{G}$. At each optimization step $k$, the estimated clear image $J_{gs}$ is generated through the rasterization of current Gaussian parameters as:
\begin{equation}
J_{gs} = \Phi(\mathcal{G}^{(k)}) \tag{9}
\end{equation}

\paragraph{the Transmission Branch ($\text{Net}_T \rightarrow t_{pred}$)}
A lightweight U-Net architecture is employed to estimate the transmission map. To prevent numerical instability such as the black hole effect where $t \rightarrow 0$ leads to division-by-zero errors, a lower-bound constraint $t_{floor}$ is introduced. The estimated transmission is expressed as:
\begin{equation}
t_{pred} = \text{Net}_T(I_{hazy}) \cdot (1 - t_{floor}) + t_{floor} \tag{10}
\end{equation}

\paragraph{the Atmospheric Light Branch($A_{grid} \rightarrow A_{map}$)} 
The atmospheric light is modeled as a spatially varying grid $A_{grid} \in \mathbb{R}^{3 \times h \times w}$ to account for non-uniform haze distribution in real-world scenes. The final full-resolution atmospheric light map $A_{map}$ is obtained via bilinear upsampling:
\begin{equation}
A_{map} = \text{Upsample}(\text{Sigmoid}(A_{logits})) \tag{11}
\end{equation}

\textbf{Physics Re-hazing and Inverse Guidance.} 
To provide supervision in the absence of paired ground-truth (GT) images, a bidirectional constraint mechanism characterized by reconstruction-inverse operations is designed. First, based on the ASM, the outputs of the three branches are recombined to generate a pseudo-hazy image $\hat{I}$:
\begin{equation}
\hat{I} = J_{gs} \odot t_{pred} + A_{map} \odot (1 - t_{pred}) \tag{12}
\end{equation}

By minimizing the reconstruction error between $\hat{I}$ and the actual input $I_{hazy}$, the parameter distributions across the three branches are constrained to adhere to physical laws. Subsequently, to guide the Gaussian field $\mathcal{G}$ toward a clear state and avoid degenerate solutions such as $J = I$ and $t = 1$, a dynamic teacher signal $J_{est}$ is constructed using the inverse physical model:
\begin{equation}
J_{est} = \frac{I_{hazy} - A_{map}(1 - t_{pred})}{\max(t_{pred}, \epsilon)} \tag{13}
\end{equation}
where $J_{est}$ represents the optimal algebraic solution under the current physical parameters. This signal serves as the optimization target to update the parameters of the Gaussian field $\mathcal{G}$ via gradient descent. To ensure the reliability of $J_{est}$, strict prior constraints are imposed on $t_{pred}$.

\begin{table}[!t]
\caption{Evaluation results of Dehaze-GaussianImage and SOTA algorithms on the NID dataset. Red and blue fonts indicate the best and second-best performance for each metric, respectively.}
\label{table_1} 
\centering
\setlength{\tabcolsep}{2pt} 
\begin{tabular}{l|ccccccc}
\toprule
Method & SSIM$\uparrow$ & PSNR$\uparrow$ & LPIPS$\downarrow$ & NIQE$\downarrow$ & LOE$\downarrow$ & DE$\uparrow$ & EME$\uparrow$ \\
\midrule
PSD \cite{ref7} & 0.81 & 15.93 & 0.12 & \textbf{\textcolor{red}{3.07}} & 118.62 & 0.15 & 11.86 \\
DehazeNet \cite{ref8} & 0.83 & 18.39 & 0.08 & 3.33 & 89.34 & 0.51 & \textbf{\textcolor{red}{20.18}} \\
Aod-Net \cite{ref9} & 0.91 & 21.53 & 0.07 & 3.34 & 96.08 & 0.35 & 4.99 \\
GridDehazeNet \cite{ref11} & 0.84 & 18.17 & 0.12 & 3.13 & 110.67 & \textbf{\textcolor{blue}{0.58}} & \textbf{\textcolor{blue}{17.12}} \\
DCP \cite{ref12} & 0.86 & 19.88 & 0.21 & 3.87 & 96.52 & 0.13 & 4.63 \\
Cycle-Dehaze \cite{ref14} & 0.79 & 14.82 & 0.26 & 4.28 & 189.63 & 0.08 & 3.63 \\
YoLY \cite{ref16} & 0.83 & 18.30 & 0.13 & 3.62 & 96.51 & 0.15 & 3.91 \\
RefineDNet \cite{ref17} & 0.90 & 19.36 & 0.11 & 3.59 & 84.28 & 0.23 & 7.87 \\
USID-Net \cite{ref18} & 0.89 & 22.56 & 0.12 & \textbf{\textcolor{blue}{3.10}} & 88.45 & 0.45 & 6.53 \\
D4 \cite{ref19} & \textbf{\textcolor{blue}{0.94}} & \textbf{\textcolor{red}{23.53}} & \textbf{\textcolor{red}{0.04}} & 3.86 & \textbf{\textcolor{red}{49.62}} & 0.39 & 5.35 \\
Diff-Dehazer \cite{ref22} & 0.76 & 11.66 & 0.20 & 4.29 & 102.68 & 0.07 & 2.55 \\
DehazeSB \cite{ref24} & 0.80 & 14.70 & 0.21 & 4.27 & 108.87 & 0.22 & 3.90 \\
\midrule 
\textbf{Ours} & \textbf{\textcolor{red}{0.95}} & \textbf{\textcolor{blue}{22.77}} & \textbf{\textcolor{blue}{0.06}} & 3.12 & \textbf{\textcolor{blue}{52.32}} & \textbf{\textcolor{red}{0.62}} & 9.00 \\
\bottomrule
\end{tabular}
\end{table}

\subsection{Adaptive Gaussian Structure Evolution}
During dehazing, the gradual removal of haze reveals previously obscured details, resulting in significant shifts in the spatial frequency characteristics of the scene. Since static geometric structures cannot accommodate these dynamic variations, a Gaussian-Aware Structural Evolution strategy driven by physical residuals is proposed. This mechanism enables Gaussian primitives to undergo adaptive splitting, cloning, pruning, and merging throughout the optimization process.

\textbf{Error-Driven Densification.} 
A physical alignment residual map is defined as $E = \|J_{gs} - J_{est}\|_1$. Regions with high values in $E$ indicate that the existing Gaussian distribution is insufficient to represent the reconstructed sharp textures. Consequently, for points with large positional gradients $\nabla_\mu \mathcal{L}$, new Gaussian primitives are cloned near their coordinates to enhance the capture of high-frequency details:
\begin{equation}
\mu_{new} = \mu_{old} + \xi, \quad 
\mu_{child} = \mu_{parent} \pm \delta \tag{14}
\end{equation}

\begin{table}[!t]
\caption{Evaluation results of Dehaze-GaussianImage and SOTA algorithms on the SOTS dataset. Red and blue fonts indicate the best and second-best performance for each metric, respectively.}
\label{table_2} 
\centering
\setlength{\tabcolsep}{2pt} 
\begin{tabular}{l|ccccccc}
\toprule
Method & SSIM$\uparrow$ & PSNR$\uparrow$ & LPIPS$\downarrow$ & NIQE$\downarrow$ & LOE$\downarrow$ & DE$\uparrow$ & EME$\uparrow$ \\
\midrule
PSD \cite{ref7} & 0.77 & 16.02 & 0.16 & 3.23 & 141.57 & 0.06 & 4.84 \\
DehazeNet \cite{ref8} & 0.57 & 12.73 & 0.45 & 4.83 & 68.69 & 0.51 & 12.92 \\
Aod-Net \cite{ref9} & 0.86 & 17.24 & 0.13 & 3.34 & \textbf{\textcolor{blue}{34.32}} & 0.34 & 7.80 \\
GridDehazeNet \cite{ref11} & \textbf{\textcolor{blue}{0.93}} & \textbf{\textcolor{red}{27.17}} & \textbf{\textcolor{red}{0.06}} & 2.70 & \textbf{\textcolor{red}{31.41}} & 0.49 & 5.30 \\
DCP \cite{ref12} & 0.79 & 14.71 & 0.18 & \textbf{\textcolor{red}{2.63}} & 55.08 & 0.50 & 8.48 \\
Cycle-Dehaze \cite{ref14} & 0.83 & 12.74 & 0.29 & 3.20 & 206.86 & 0.47 & 3.54 \\
YoLY \cite{ref16} & 0.53 & 19.08 & 0.22 & 2.98 & 135.12 & 0.11 & 2.87 \\
RefineDNet \cite{ref17} & 0.90 & 20.25 & 0.13 & 3.10 & 85.06 & 0.35 & 5.48 \\
USID-Net \cite{ref18} & 0.75 & 21.40 & 0.17 & 3.91 & 55.33 & 0.41 & 5.38 \\
D4 \cite{ref19} & 0.83 & 18.68 & \textbf{\textcolor{blue}{0.07}} & \textbf{\textcolor{blue}{2.49}} & 56.48 & 0.46 & \textbf{\textcolor{blue}{13.63}} \\
Diff-Dehazer \cite{ref22} & 0.91 & 23.38 & 0.11 & 2.69 & 41.16 & 0.51 & 4.16 \\
DehazeSB \cite{ref24} & 0.84 & 21.74 & 0.17 & 2.80 & 121.22 & \textbf{\textcolor{red}{0.62}} & 7.03 \\
\midrule
\textbf{Ours} & \textbf{\textcolor{red}{0.94}} & \textbf{\textcolor{blue}{25.07}} & 0.14 & 3.09 & 74.37 & \textbf{\textcolor{blue}{0.61}} & \textbf{\textcolor{red}{16.40}} \\
\bottomrule
\end{tabular}
\end{table}

To maintain representational compactness, primitives with minimal visual contribution are removed at predefined iteration intervals. The pruning criterion is formulated based on opacity $\alpha$ and geometric scale $r$:
\begin{equation}
\mathcal{M}_{prune} = \{i \mid \alpha_i < \tau_\alpha \vee r_i > \tau_{radius}\} \tag{15}
\end{equation}

To mitigate artifacts resulting from the over-accumulation of Gaussian primitives in homogeneous regions, a grid-based merging strategy is implemented. Within each spatial grid cell $\Omega_k$, overlapping Gaussian primitives are fused into a single primitive via opacity-weighted averaging:
\begin{equation}
\mu_{merge} = \frac{\sum_{j \in \Omega_k} \alpha_j \mu_j}{\sum \alpha_j}, \quad 
c_{merge} = \frac{\sum_{j \in \Omega_k} \alpha_j c_j}{\sum \alpha_j} \tag{16}
\end{equation}

This dynamic evolution mechanism ensures that the Gaussian field maintains an optimal and compact geometric configuration. Such a distribution facilitates the concurrent optimization of geometric structures and physical attributes during the dehazing process.

\subsection{Physics-Constrained Loss Functions}
To mitigate the uncertainty inherent in unsupervised training and enhance visual quality, a composite loss function framework is formulated. This framework integrates self-supervised reconstruction, physical prior constraints, and geometric regularization to provide robust supervision.

\textbf{Self-Supervised Reconstruction Loss.} 
This loss serves as the primary objective during the second stage. By leveraging the decoupled components $J_{gs}$, $t_{pred}$, and $A_{map}$, a pseudo-hazy image $\hat{I}$ is synthesized based on the atmospheric scattering model and constrained to approximate the observed hazy image $I_{hazy}$. This formulation enforces the mathematical validity of the decomposed physical parameters:
\begin{equation}
\mathcal{L}_{recon} = \lambda_{L1} \|\hat{I} - I_{hazy}\|_1 + \lambda_{ssim} (1 - \text{SSIM}(\hat{I}, I_{hazy})) \tag{17}
\end{equation}
where a combination of the $L_1$ norm and SSIM is employed in the RGB space to ensure color fidelity and structural consistency simultaneously.

\begin{table*}[!t]
\caption{Evaluation results of Dehaze-GaussianImage and SOTA algorithms on the RTTS and Haze2020 datasets. Red and blue fonts indicate the best and second-best performance for each metric, respectively.}
\label{table_3} 
\centering
\begin{tabular}{l|cccc|cccc}
\toprule
\multirow{2}{*}{Method} & \multicolumn{4}{c|}{RTTS} & \multicolumn{4}{c}{Haze2020} \\
 & FID$\downarrow$ & NIQE$\downarrow$ & MUSIQ$\uparrow$ & MANIQA$\uparrow$ & FID$\downarrow$ & NIQE$\downarrow$ & MUSIQ$\uparrow$ & MANIQA$\uparrow$ \\
\midrule
PSD \cite{ref7} & 65.3 & 4.04 & 50.1 & 0.14 & 79.5 & 3.92 & 55.0 & 0.19 \\
DehazeNet \cite{ref8} & 79.3 & 4.75 & 47.2 & 0.11 & 89.4 & 4.07 & 52.9 & 0.09 \\
Aod-Net \cite{ref9} & 73.8 & 3.92 & 51.3 & 0.12 & 85.3 & 3.86 & 54.2 & 0.15 \\
GridDehazeNet \cite{ref11} & 69.3 & 3.77 & 49.7 & 0.17 & 80.2 & 3.95 & 53.7 & 0.15 \\
DCP \cite{ref12} & 74.2 & 4.31 & 50.4 & 0.12 & 81.7 & 3.81 & 53.4 & 0.12 \\
Cycle-Dehaze \cite{ref14} & 88.7 & 4.92 & 46.7 & 0.11 & 91.3 & 4.27 & 52.3 & 0.11 \\
YoLY \cite{ref16} & 71.3 & 4.55 & 52.3 & 0.13 & 84.5 & 3.88 & 54.7 & 0.13 \\
RefineDNet \cite{ref17} & 64.2 & 4.21 & 56.3 & 0.17 & 87.3 & 3.98 & 55.3 & 0.11 \\
USID-Net \cite{ref18} & 61.8 & 4.10 & 57.7 & 0.14 & 85.2 & 3.96 & 53.2 & 0.15 \\
D4 \cite{ref19} & 70.5 & 3.96 & 58.3 & 0.15 & 84.5 & 3.91 & 53.5 & 0.17 \\
Diff-Dehazer \cite{ref22} & \textbf{\textcolor{red}{52.3}} & 3.84 & \textbf{\textcolor{blue}{59.4}} & \textbf{\textcolor{red}{0.33}} & 70.9 & \textbf{\textcolor{blue}{3.57}} & \textbf{\textcolor{red}{61.3}} & \textbf{\textcolor{red}{0.38}} \\
DehazeSB \cite{ref24} & \textbf{\textcolor{blue}{53.1}} & \textbf{\textcolor{blue}{3.74}} & \textbf{\textcolor{red}{61.3}} & 0.17 & \textbf{\textcolor{blue}{69.8}} & 3.64 & \textbf{\textcolor{blue}{59.7}} & 0.18 \\
\midrule
\textbf{Ours} & 57.2 & \textbf{\textcolor{red}{3.01}} & 53.6 & \textbf{\textcolor{blue}{0.19}} & \textbf{\textcolor{red}{68.9}} & \textbf{\textcolor{red}{3.44}} & 56.6 & \textbf{\textcolor{blue}{0.31}} \\
\bottomrule
\end{tabular}
\end{table*}

\textbf{Transmission Guidance with Brightness Protection.} 
The application of the traditional DCP often leads to the erroneous estimation of sky regions or luminous objects as dense distant haze ($t \approx 0$). To mitigate this intrinsic limitation, a brightness-protected guidance loss is formulated. Initially, a baseline DCP guidance map is computed:
\begin{equation}
t_{base}(x) = 1 - \omega \cdot \min_{c \in \{r,g,b\}} \left( \min_{y \in \Omega(x)} \frac{I_{hazy}(y)}{A_{map}(y)} \right) \tag{18}
\end{equation}

Subsequently, a brightness protection mask $M_{bright}$ is constructed to distinguish high-intensity regions such as clouds. For these areas, the transmission is constrained to approach unity:
\begin{equation}
\begin{cases} 
M_{bright}(x) = \text{Clamp}\left(\dfrac{\max_c I(x) - \tau_{high}}{1 - \tau_{high}}, 0, 1\right) \\[10pt]
t_{guide}(x) = t_{base}(x) \cdot (1 - M_{bright}(x)) \\
\qquad\qquad\quad + 1.0 \cdot M_{bright}(x)
\end{cases} \tag{19}
\end{equation}

The final guidance loss is expressed as:
\begin{equation}
\mathcal{L}_{guide} = \|t_{pred} - t_{guide}\|_2^2 \tag{20}
\end{equation}

\textbf{Dark Channel Hinge Loss for $J_{est}$.} 
To further regulate the dehazing level of the estimated latent image $J_{est}$, a DCP constraint is imposed. A hinge loss formulation is adopted to prevent chromatic attenuation resulting from excessive dehazing. This loss effectively eliminates interference from sky regions $M_{sky}$ via a weighting factor $\omega(x)$:
\begin{equation}
\mathcal{L}_{dcp} = \frac{1}{\sum \omega} \sum_x \omega(x) \cdot \text{ReLU}(\text{DC}(J_{est}(x)) - \tau_{dc}) \tag{21}
\end{equation}
where $\omega(x) = 1 - M_{sky}(x)$ denotes the spatial weight.

\textbf{Gray Retention Loss.} 
To mitigate the undesirable darkening of light gray objects such as roads or concrete during the dehazing process, a gray-scale retention loss is formulated. In non-sky regions, the radiance of the latent clear image $J$ is expected to maintain a specific proportion relative to the original image $I$. To this end, a weight mask $W_{gray}$ is designed to prioritize pixels characterized by low saturation and moderate luminance:
\begin{equation}
\begin{aligned}
W_{gray} = & I(S(I) < 0.3) \cdot I(0.3 < L(I) < 0.9) \\
& \cdot (1 - M_{sky})
\end{aligned} \tag{22}
\end{equation}

The loss function penalizes excessive brightness attenuation and is expressed as follows:
\begin{equation}
\begin{aligned}
\mathcal{L}_{gray} = & \sum_x W_{gray}(x) \cdot \text{ReLU}((\max_c I(x) \cdot \gamma) \\
& - \max_c J_{est}(x)) 
\end{aligned}\tag{23}
\end{equation}

Specifically, $\gamma \approx 0.7$ is utilized as the brightness retention factor.

\textbf{Gaussian-Physical Alignment Loss.} 
To distill the dehazing knowledge acquired by the physical branches into the Gaussian field, we define an alignment loss. Considering that regions with dense haze (low $t$) involve high restoration uncertainty, we utilize $t$ as confidence weights:
\begin{equation}
\mathcal{L}_{align} = \sum_x (t_{pred}(x))^\lambda \cdot \|J_{gs}(x) - sg(J_{est}(x))\|_1 \tag{24}
\end{equation}
where $sg(\cdot)$ denotes the stop-gradient operation, which prevents the optimization of the Gaussian field from backward-interfering with the learning of physical parameters.

\textbf{Regularization Terms.} 
To enhance perceptual quality, several regularization terms are incorporated into the optimization objective. Specifically, to maintain the spatial smoothness of the transmission map while preserving depth discontinuities, an edge-weighted Total Variation (TV) loss $\mathcal{L}_{tv}$ is formulated:
\begin{equation}
\mathcal{L}_{tv} = \sum_x e^{-\mathcal{K} |\nabla I|} \cdot (|\nabla_x t_{pred}| + |\nabla_y t_{pred}|) \tag{25}
\end{equation}

Additionally, to suppress color noise and minimize variance within sky regions, a Sky Uniformity loss $\mathcal{L}_{sky}$ is applied to the sky-masked areas:
\begin{equation}
\mathcal{L}_{sky} = \text{Var}(J_{gs}(x) \mid x \in M_{sky}) \tag{26}
\end{equation}

Furthermore, to mitigate unnatural spherical artifacts with excessive radii in homogeneous regions, a Blob Suppression loss $\mathcal{L}_{blob}$ is introduced. This loss leverages a smoothness-aware operator $M_{smooth}$ to penalize Gaussian primitives that exceed a predefined radius threshold $r_{th}$:
\begin{equation}
\mathcal{L}_{blob} = \sum_i \alpha_i \cdot \text{Sigmoid}(r_i - r_{th}) \cdot M_{smooth}(\boldsymbol{\mu}_i) \tag{27}
\end{equation}

Lastly, to suppress needle-like artifacts induced by the extreme elongation of the covariance matrix, an Anisotropy Regularization term $\mathcal{L}_{aniso}$ is introduced:
\begin{equation}
\mathcal{L}_{aniso} = \sum_i \text{ReLU}\left(\frac{\lambda_{max}(\Sigma_i)}{\lambda_{min}(\Sigma_i)} - \tau_{aniso}\right) \tag{28}
\end{equation}

The aforementioned geometric constraints are ultimately consolidated into a unified regularization term $\mathcal{L}_{regs}$, formulated as a weighted combination of the constituent loss components:
\begin{equation}
\mathcal{L}_{regs} = \lambda_{tv} \mathcal{L}_{tv} + \lambda_{sky} \mathcal{L}_{sky} + \lambda_{aniso} \mathcal{L}_{aniso} + \lambda_{blob} \mathcal{L}_{blob} \tag{29}
\end{equation}

\textbf{Total Objective Function.} 
Ultimately, the Dehaze-GaussianImage framework is optimized in an end-to-end manner by minimizing the global objective function $\mathcal{L}_{total}$. This formulation integrates multiple constraints spanning from pixel-level reconstruction to high-level physical priors:
\begin{equation}
\begin{aligned}
\mathcal{L}_{total} = & \lambda_{recon}\mathcal{L}_{recon} + \lambda_{align}\mathcal{L}_{align} + \lambda_{guide}\mathcal{L}_{guide} + \\
& \lambda_{dcp}\mathcal{L}_{dcp} + \lambda_{gray}\mathcal{L}_{gray} + \mathcal{L}_{regs}    
\end{aligned} \tag{30} 
\end{equation}

The weighting coefficients $\lambda$ for each component are specified in the experimental section. Through the joint optimization of this objective function, high-quality dehazing and scene disentanglement are achieved without the requirement for paired training data.

\begin{table}[!t]
\centering
\caption{Ablation study of different loss functions.}
\label{table_4} 
\setlength{\tabcolsep}{2pt} 
\begin{tabular}{ccccc|ccc}
\toprule
$\mathcal{L}_{recon}+\mathcal{L}_{dcp}$ & $\mathcal{L}_{ssim}$ & $\mathcal{L}_{fft}$ & $\mathcal{L}_{edge}$ & $\mathcal{L}_{grad}$ & PSNR$\uparrow$ & SSIM$\uparrow$ & LPIPS$\downarrow$ \\
\midrule
\checkmark & ×   & ×   & ×   & ×   & 21.32 & 0.88 & 0.17 \\
\checkmark & \checkmark & ×   & ×   & ×   & 22.71 & 0.91 & 0.12 \\
\checkmark & \checkmark & \checkmark & ×   & ×   & 22.90 & 0.95 & 0.12 \\
\checkmark & \checkmark & \checkmark & \checkmark & ×   & \textbf{\textcolor{blue}{23.72}} & \textbf{\textcolor{blue}{0.96}} & \textbf{\textcolor{blue}{0.08}} \\
\checkmark & \checkmark & \checkmark & \checkmark & \checkmark & \textbf{\textcolor{red}{24.31}} & \textbf{\textcolor{red}{0.96}} & \textbf{\textcolor{red}{0.04}} \\
\bottomrule
\end{tabular}
\end{table}

\begin{table}[!t]
\caption{Ablation study of different evolution strategies. Red and blue fonts indicate the best and second-best performance for each metric, respectively.}
\label{table_5} 
\centering
\begin{tabular}{l|ccc}
\toprule
Evolution Strategy & PSNR$\uparrow$ & SSIM$\uparrow$ & LPIPS$\downarrow$ \\
\midrule
Fixed Topology & \textbf{\textcolor{blue}{21.32}} & \textbf{\textcolor{blue}{0.85}} & 0.19 \\
Prune Only & 19.24 & 0.76 & \textbf{\textcolor{blue}{0.13}} \\
Densify \& Split Only & 18.42 & 0.71 & 0.21 \\
\midrule
\textbf{Full Dynamics (Ours)} & \textbf{\textcolor{red}{24.31}} & \textbf{\textcolor{red}{0.96}} & \textbf{\textcolor{red}{0.04}} \\
\bottomrule
\end{tabular}
\end{table}

\section{Experiments}
\subsection{Experimental Setup}
\textbf{Datasets.} 
To evaluate the performance of Dehaze-GaussianImage regarding objective physical consistency and subjective visual perception, four representative dehazing datasets are utilized. Samples are partitioned into training and testing sets following a 9:1 ratio. These datasets are classified into two categories. The first category includes the SOTS~\cite{ref51} and NID~\cite{ref52} datasets. Specifically, SOTS contains indoor and outdoor scenes, while NID encompasses light and dense haze levels. Since these datasets provide paired hazy and ground-truth images, they are used for full-reference evaluation to quantify the accuracy of pixel-level reconstruction and structural fidelity. The second category consists of the RTTS~\cite{ref53} and Haze2020~\cite{ref54} datasets, which comprise complex real-world hazy scenes without reference images. Consequently, these datasets are employed for no-reference evaluation to assess the generalization capability and visual naturalness of the proposed method in authentic degradation scenarios.

\textbf{Implementation Details.} 
The proposed method is implemented within the PyTorch framework, and all experiments are conducted on a single NVIDIA RTX 3090 GPU. The reconstruction-decoupling optimization process consists of two stages. In the first stage, the Gaussian field for the hazy image is initialized with $10^4$ 2D Gaussian primitives. The reconstruction is performed for $3 \times 10^4$ iterations using the Adam optimizer with a base learning rate of 0.01. The second stage involves zero-shot physical dehazing totaling 9,000 iterations. This stage includes 2,000 warm-up iterations for the transmission network and 7,000 iterations for joint Gaussian optimization. Specifically, the transmission map is estimated via a lightweight U-Net, where parameters are updated in coordination with the 2D Gaussian attributes. The learning rates for the network and Gaussian colors are set to $6 \times 10^{-4}$ and $5 \times 10^{-4}$, respectively. To improve representation efficiency and capture structural details, a dynamic density control strategy is integrated into the Gaussian parameter space. During optimization, Gaussian primitives undergo adaptive densification and splitting. Redundant points with opacity values below 0.02 are pruned every 200 steps. Furthermore, the total number of Gaussian primitives is constrained to a maximum of $10^5$ to ensure high-quality dehazing with minimal memory usage and computational cost.

\textbf{Evaluation Metrics.} 
Two distinct evaluation systems are implemented to accommodate various data characteristics. For the SOTS and NID datasets which contain paired reference images, seven metrics are selected for comprehensive assessment. These include three full-reference metrics, PSNR, SSIM, and LPIPS, along with four no-reference metrics consisting of NIQE, LOE, DE, and EME. Full-reference metrics are utilized to ensure consistency between the dehazed and ground-truth images. Specifically, PSNR measures pixel-level fidelity, SSIM assesses similarity in terms of luminance, contrast, and structure, and LPIPS evaluates perceptual distance within the VGG-based feature space. Concurrently, no-reference metrics are employed to quantify naturalness, contrast, and information content. Among these, NIQE measures the deviation from natural image statistics, LOE evaluates the preservation of natural illumination, DE indicates the richness of information, and EME quantifies local contrast variations. 

Furthermore, for real-world scenarios in the RTTS and Haze2020 datasets where reference images are unavailable, four advanced no-reference metrics are adopted, including FID, NIQE, MUSIQ, and MANIQA. Specifically, FID calculates the Fréchet distance between the feature distributions of generated and real images. MUSIQ utilizes a multi-scale Transformer to capture quality across different scales, while MANIQA leverages a multi-dimensional attention mechanism for global and local perception. These metrics collectively reflect the visual aesthetic performance of the proposed method in complex environments.

\subsection{Performance Comparison}
To verify the effectiveness of the proposed Dehaze-GaussianImage, its performance is compared with four supervised SOTA methods~\cite{ref7,ref8,ref9,ref11} and eight unsupervised SOTA methods~\cite{ref12,ref14,ref16,ref17,ref18,ref19,ref22,ref24}. The quantitative results on four datasets are reported in Tables~\ref{table_1}--\ref{table_3}.

\textbf{Visual Comparison.} 
Fig.~\ref{fig_3} illustrates the qualitative results on the NID dataset. Regarding the overall haze removal, existing methods such as AOD-Net, Cycle-Dehaze, YoLY, D4, DCP, and Diff-Dehazer exhibit significant residual haze or non-uniform dehazing artifacts. In contrast, Dehaze-GaussianImage achieves superior performance in both haze removal intensity and color fidelity. For sharpness, DehazeNet and GridDehazeNet produce excessively high contrast. Conversely, USID-Net and the proposed algorithm restore clear features while maintaining moderate transparency and natural visual effects. Furthermore, USID-Net and the proposed method demonstrate the best comprehensive performance in color representation.

The visual comparison on the SOTS dataset is presented in Fig.~\ref{fig_4}. Specifically, DCP, PSD, Cycle-Dehaze, and DehazeNet show significant color distortion. Meanwhile, AOD-Net, GridDehazeNet, D4, and RefineDNet suffer from excessive contrast and insufficient brightness. In terms of overall clarity, DehazeSB and the proposed algorithm achieve the best results. In contrast, several methods, including Diff-Dehazer and YoLY, fail to remove haze completely, leaving noticeable local artifacts.

\textbf{Detailed Restoration Evaluation.} 
To further demonstrate the superiority of Dehaze-GaussianImage in restoring fine textures, complex scenarios from the SOTS dataset are selected for visual verification. As shown in Fig.~\ref{fig_5}, Dehaze-GaussianImage aligns closely with the ground-truth images regarding structural preservation, color saturation, and contrast control. Conversely, other methods suffer from localized color distortion or excessive contrast, which indicates deficiencies in visual authenticity and detail fidelity.

Overall, Dehaze-GaussianImage outperforms existing SOTA methods in both global appearance and local detail reconstruction. Furthermore, the proposed framework exhibits superior capabilities in feature recovery and haze removal compared to current mainstream algorithms, showcasing its significant comprehensive advantages.

\textbf{Quantitative Evaluation.} 
To evaluate the performance on the paired NID and SOTS datasets, full-reference metrics (PSNR and SSIM) and no-reference metrics (NIQE, LOE, DE, and EME) are adopted. Tables~\ref{table_1} and~\ref{table_2} summarize the results, showing that Dehaze-GaussianImage achieves superior performance on both datasets. Specifically, the proposed method consistently ranks within the top two for the core metrics PSNR and SSIM. Moreover, its performance across other evaluation metrics remains competitive, leading the overall ranking. On the two unpaired datasets, Table~\ref{table_3} demonstrates that Dehaze-GaussianImage ranks among the top three in both FID and NIQE while maintaining robust performance across other metrics. Overall, the proposed compressed-domain dehazing paradigm exhibits superior performance across diverse quantitative evaluations. These results highlight the effectiveness and innovation of our approach compared to existing SOTA methods.

\subsection{Ablation Study}
To evaluate the contribution of each core component in the proposed Dehaze-GaussianImage framework, ablation studies are conducted on the NID dataset. PSNR, SSIM, and LPIPS are adopted as quantitative metrics, and qualitative comparisons of local regions are provided to demonstrate the effectiveness of our approach. These evaluations confirm the superior performance in structural fidelity, color restoration, and physical parameter decoupling.

\textbf{Comprehensive Ablation on Loss Functions.} 
Several specialized loss functions are integrated into the unsupervised physical decoupling process to handle the limitations of traditional priors in extreme regions and regularize the 2DGS morphology. The baseline configuration includes the basic reconstruction loss $\mathcal{L}_{recon}$ and the standard dark channel prior $\mathcal{L}_{dcp}$. Subsequently, the proposed key modules are progressively added to evaluate their individual contributions. 

As shown in Table~\ref{table_4}, each component is essential to the framework. When all loss functions are employed, Dehaze-GaussianImage achieves a 2.99 dB improvement in PSNR over the baseline, which validates the effectiveness of the proposed loss design.

\textbf{Impact of Dynamic Gaussian Morphological Evolution.} 
Dehazing fundamentally involves the gradual recovery of high-frequency textures. In this work, the dynamic evolution of 2DGS is reactivated during the decoupling stage. To evaluate the proposed architecture, four morphological evolution strategies are compared. (a) Fixed Topology: the quantity and positions of Gaussian primitives are fixed, and only color and opacity are optimized. (b) Prune Only: only the pruning mechanism is activated to remove haze-affected primitives. (c) Densify \& Split Only: only error-driven densification and splitting are enabled. (d) Full Dynamics: the complete adaptive evolution mechanism is employed. 

As reported in Table~\ref{table_5}, the Full Dynamics mechanism combines transparency pruning with periodic Gaussian merging. This mechanism captures high-frequency structures adaptively while maintaining an efficient point cloud scale, achieving the optimal balance between fidelity and representation efficiency.

\textbf{Spatial Modeling of Atmospheric Light.} 
Atmospheric light is conventionally assumed to be a globally uniform RGB vector in physical dehazing methods. To handle non-uniform illumination and inhomogeneous haze distribution, the proposed framework models atmospheric light as an optimizable low-resolution spatial grid ($16 \times 16$). Three approaches are evaluated for comparison. 1) Global Static Vector: a global vector fixed after dark channel calculation. 2) Global Optimized Vector: a global vector treated as a learnable parameter. 3) Spatial Grid ($16 \times 16$): the proposed spatial optimization strategy. 

As shown in Table~6, the Spatial Grid strategy fits the non-uniform scattered light adaptively. When combined with a bilinear upsampling mechanism, this strategy enables smooth and localized estimation of atmospheric light across the image. Consequently, substantial improvements are achieved in PSNR and SSIM.

\subsection{Limitation and Future works}
Despite its superior performance in unsupervised single-image dehazing, Dehaze-GaussianImage requires independent iterative optimization for each image. This leads to substantial inference latency and limits its suitability for real-time applications. Additionally, physical priors are susceptible to performance degradation in scenarios involving extreme haze or complex nighttime illumination. Future efforts will focus on incorporating feed-forward networks to predict initial 2DGS parameters for enhanced generalization and speed. Furthermore, the extension of this framework to video sequences and non-uniform nighttime dehazing tasks will be investigated.

\section{Conclusion}
In this paper, we propose Dehaze-GaussianImage, the first zero-shot single image dehazing framework based on 2DGS. Departing from the conventional pixel-level processing paradigm, a reconstruction-decoupling strategy is introduced to integrate the atmospheric scattering model into the Gaussian parameter space. Employing the adaptive dynamic evolution of Gaussian primitives, the proposed method achieves geometric decoupling between clear textures and the hazy medium. Explicit structure-preserving constraints are incorporated to eliminate artifacts typically associated with traditional physical priors. Experimental results demonstrate that the proposed framework outperforms SOTA methods in both dehazing fidelity and visual quality under completely unsupervised conditions while maintaining a minimal parameter count. This work establishes an efficient and physically interpretable paradigm for image dehazing. Furthermore, it highlights the significant potential of explicit 2DGS representations in low-level computer vision tasks.

\begin{IEEEbiography}[{\includegraphics[width=1in,height=1.25in,clip,keepaspectratio]{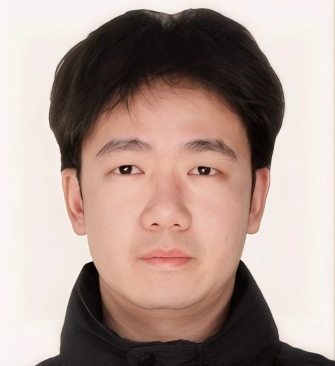}}]{Yuhan Chen}
received his master's degree in 2024 from the College of Mechanical Engineering at Chongqing University of Technology. He is currently pursuing the Ph.D. degree in College of Mechanical and Vehicle Engineering at Chongqing University, China. His research interests include deep learning, Low-level Vision and Gaussian Splatting.
\end{IEEEbiography}
\vspace{-40pt}

\begin{IEEEbiography}[{\includegraphics[width=1in,height=1.25in,clip,keepaspectratio]{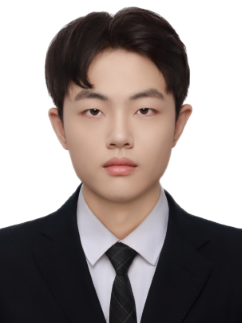}}]{Wenxuan Yu}
received the B.E. degree majoring in Mechanical Design, Manufacturing, and Automation at Chongqing University in 2025. He is currently pursuing the M.E. degree in Mechanical Engineering at Chongqing University, Chongqing, China. His research interests include computer vision, Gaussian Splatting and deep learning.
\end{IEEEbiography}
\vspace{-30pt}

\linespread{0.9}\selectfont
\begin{IEEEbiography}[{\includegraphics[width=1in,height=1.25in,clip,keepaspectratio]{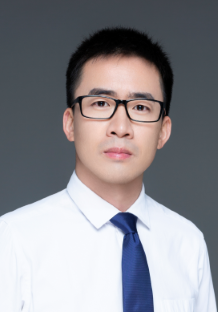}}]{Guofa Li}
received the Ph.D. degree in Mechanical Engineering from Tsinghua University, China, in 2016. He is currently a Professor with Chongqing University, China. His research interests include environment perception, driver behavior analysis, and smart decision-making based on artificial intelligence technologies in autonomous vehicles and intelligent transportation systems. He serves as the Associate Editor for IEEE Transactions on Intelligent Transportation Systems, IEEE Transactions on Affective Computing, and IEEE Sensors Journal.
\end{IEEEbiography}
\vspace{-40pt}

\begin{IEEEbiography}[{\includegraphics[width=1in,height=1.25in,clip,keepaspectratio]{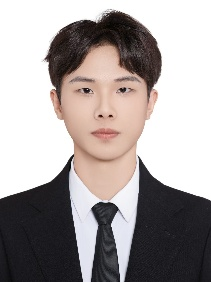}}]{Ying Fang}
received the B.E. degree majoring in Vehicle Engineering at Chongqing University of Technology. He is currently pursuing the M.E. degree in Mechanical Engineering at Chongqing University, Chongqing, China. His research interests include computer vision, Gaussian Splatting and deep learning.
\end{IEEEbiography}

\begin{IEEEbiography}[{\includegraphics[width=1in,height=1.25in,clip,keepaspectratio]{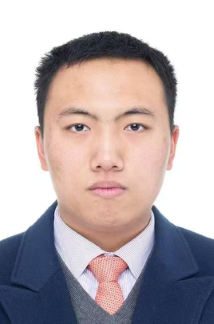}}]{Kunyang Huang}
is currently an masters degree student at Carnegie Mellon University. He previously interned at the Chongqing Institute of the Chinese Academy of Sciences and a research intern at the Changan Automobile Research Institute. His research interests include autonomous driving perception and deep learning methods.
\end{IEEEbiography}
\vspace{-130pt}

\vspace{-160pt}
\begin{IEEEbiography}[{\includegraphics[width=1in,height=1.25in,clip,keepaspectratio]{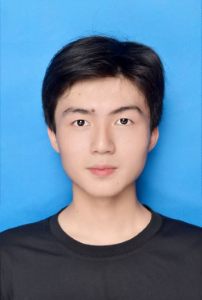}}]{Yicui Shi}
received the B.E. degree majoring in Automotive Engineering at Chongqing University in 2025. He is currently pursuing the M.E. degree in Automotive Engineering at Chongqing University, Chongqing, China. His research interests include computer vision, Gaussian Splatting and deep learning.
\end{IEEEbiography}
\vspace{-150pt}

\vspace{-140pt}
\linespread{0.9}\selectfont
\begin{IEEEbiography}[{\includegraphics[width=1in,height=1.25in,clip,keepaspectratio]{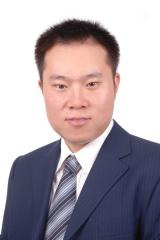}}]{Wenbo Chu}
received his B.S. degree majored in Automotive Engineering from Tsinghua University, China, in 2008, and his M.S. degree majored in Automotive Engineering from RWTH-Aachen, German and Ph.D. degree majored in Mechanical Engineering from Tsinghua University, China, in 2014. He is currently a research fellow at Western China Science City Innovation Center of Intelligent and Connected Vehicles (Chongqing) Co, Ltd., and National Innovation Center of Intelligent and Connected Vehicles.
\end{IEEEbiography}
\vspace{-150pt}

\vspace{-140pt}
\linespread{0.9}\selectfont
\begin{IEEEbiography}[{\includegraphics[width=1in,height=1.25in,clip,keepaspectratio]{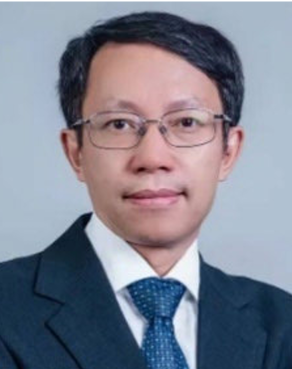}}]{Keqiang Li}
received the B.E. degree from Tsinghua University, Beijing, China, in 1985, and the M.E. and Ph.D. degrees from Chongqing University, Chongqing, China, in 1988 and 1995, respectively. He is currently a Professor with the School of Vehicle and Mobility, Tsinghua University. He is the Chief Scientist of Intelligent and Connected Vehicle Innovation Center of China, and the Director of State Key Laboratory of Automotive Safety and Energy of China. His current research interests include intelligent connected vehicles, cloud-based control for vehicles, and vehicle dynamics systems.
\end{IEEEbiography}


\begin{thebibliography}{99}
\bibliographystyle{IEEEtran}

\bibitem{ref1}
A. Ayoub, W. El-Shafai, F. E. A. El-Samie, et al., ``Review of dehazing techniques: Challenges and future trends,'' \textit{Multimedia Tools Appl.}, vol. 84, no. 3, pp. 1103--1131, 2025.

\bibitem{ref2}
C. Li, W. Yan, H. Zhao, et al., ``TFFD-Net: An effective two-stage mixed feature fusion and detail recovery dehazing network,'' \textit{Vis. Comput.}, vol. 41, no. 6, pp. 4001--4016, 2025.

\bibitem{ref3}
S. C. Agrawal and A. S. Jalal, ``A comprehensive review on analysis and implementation of recent image dehazing methods,'' \textit{Arch. Comput. Methods Eng.}, vol. 29, no. 7, pp. 4799--4850, 2022.

\bibitem{ref4}
Y. Qu, Y. Chen, J. Huang, et al., ``Enhanced pix2pix dehazing network,'' in \textit{Proc. IEEE/CVF Conf. Comput. Vis. Pattern Recognit. (CVPR)}, 2019, pp. 8160--8168.

\bibitem{ref5}
X. Qin, Z. Wang, Y. Bai, et al., ``FFA-Net: Feature fusion attention network for single image dehazing,'' in \textit{Proc. AAAI Conf. Artif. Intell.}, vol. 34, no. 7, pp. 11908--11915, 2020.

\bibitem{ref6}
Q. Deng, Z. Huang, C. C. Tsai, et al., ``Hardgan: A haze-aware representation distillation gan for single image dehazing,'' in \textit{Proc. Eur. Conf. Comput. Vis. (ECCV)}, Cham, Switzerland: Springer, 2020, pp. 722--738.

\bibitem{ref7}
Z. Chen, Y. Wang, Y. Yang, et al., ``PSD: Principled synthetic-to-real dehazing guided by physical priors,'' in \textit{Proc. IEEE/CVF Conf. Comput. Vis. Pattern Recognit. (CVPR)}, 2021, pp. 7180--7189.

\bibitem{ref8}
B. Cai, X. Xu, K. Jia, et al., ``DehazeNet: An end-to-end system for single image haze removal,'' \textit{IEEE Trans. Image Process.}, vol. 25, no. 11, pp. 5187--5198, 2016.

\bibitem{ref9}
B. Li, X. Peng, Z. Wang, et al., ``AOD-Net: All-in-one dehazing network,'' in \textit{Proc. IEEE Int. Conf. Comput. Vis. (ICCV)}, 2017, pp. 4770--4778.

\bibitem{ref10}
W. Ren, S. Liu, H. Zhang, et al., ``Single image dehazing via multi-scale convolutional neural networks,'' in \textit{Proc. Eur. Conf. Comput. Vis. (ECCV)}, Cham, Switzerland: Springer, 2016, pp. 154--169.

\bibitem{ref11}
X. Liu, Y. Ma, Z. Shi, et al., ``GridDehazeNet: Attention-based multi-scale network for image dehazing,'' in \textit{Proc. IEEE/CVF Int. Conf. Comput. Vis. (ICCV)}, 2019, pp. 7314--7323.

\bibitem{ref12}
K. He, J. Sun, and X. Tang, ``Single image haze removal using dark channel prior,'' \textit{IEEE Trans. Pattern Anal. Mach. Intell.}, vol. 33, no. 12, pp. 2341--2353, 2010.

\bibitem{ref13}
J. Y. Zhu, T. Park, P. Isola, et al., ``Unpaired image-to-image translation using cycle-consistent adversarial networks,'' in \textit{Proc. IEEE Int. Conf. Comput. Vis. (ICCV)}, 2017, pp. 2223--2232.

\bibitem{ref14}
D. Engin, A. Gen\c{c}, and H. Kemal Ekenel, ``Cycle-dehaze: Enhanced CycleGAN for single image dehazing,'' in \textit{Proc. IEEE Conf. Comput. Vis. Pattern Recognit. Workshops (CVPRW)}, 2018, pp. 825--833.

\bibitem{ref15}
X. Yang, Z. Xu, and J. Luo, ``Towards perceptual image dehazing by physics-based disentanglement and adversarial training,'' in \textit{Proc. AAAI Conf. Artif. Intell.}, vol. 32, no. 1, 2018.

\bibitem{ref16}
B. Li, Y. Gou, S. Gu, et al., ``You only look yourself: Unsupervised and untrained single image dehazing neural network,'' \textit{Int. J. Comput. Vis.}, vol. 129, no. 5, pp. 1754--1767, 2021.

\bibitem{ref17}
S. Zhao, L. Zhang, Y. Shen, et al., ``RefineDNet: A weakly supervised refinement framework for single image dehazing,'' \textit{IEEE Trans. Image Process.}, vol. 30, pp. 3391--3404, 2021.

\bibitem{ref18}
J. Li, Y. Li, L. Zhuo, et al., ``USID-Net: Unsupervised single image dehazing network via disentangled representations,'' \textit{IEEE Trans. Multimedia}, vol. 25, pp. 3587--3601, 2022.

\bibitem{ref19}
Y. Yang, C. Wang, R. Liu, et al., ``Self-augmented unpaired image dehazing via density and depth decomposition,'' in \textit{Proc. IEEE/CVF Conf. Comput. Vis. Pattern Recognit. (CVPR)}, 2022, pp. 2037--2046.

\bibitem{ref20}
T. Park, A. A. Efros, R. Zhang, et al., ``Contrastive learning for unpaired image-to-image translation,'' in \textit{Proc. Eur. Conf. Comput. Vis. (ECCV)}, Cham, Switzerland: Springer, 2020, pp. 319--345.

\bibitem{ref21}
Y. Wang, X. Yan, F. L. Wang, et al., ``UCL-dehaze: Toward real-world image dehazing via unsupervised contrastive learning,'' \textit{IEEE Trans. Image Process.}, vol. 33, pp. 1361--1374, 2024.

\bibitem{ref22}
Y. Lan, Z. Cui, C. Liu, et al., ``Exploiting diffusion prior for real-world image dehazing with unpaired training,'' in \textit{Proc. AAAI Conf. Artif. Intell.}, vol. 39, no. 4, pp. 4455--4463, 2025.

\bibitem{ref23}
C. Liu, L. Qi, J. Pan, et al., ``Frequency domain-based diffusion model for unpaired image dehazing,'' 2025, arXiv: 2507.01275.

\bibitem{ref24}
Y. Lan, Z. Cui, X. Luo, et al., ``When Schrodinger bridge meets real-world image dehazing with unpaired training,'' in \textit{Proc. IEEE/CVF Int. Conf. Comput. Vis. (ICCV)}, 2025, pp. 8756--8765.

\bibitem{ref25}
R. Wu, X. Wang, P. Liang, et al., ``Toward zero-shot learning for visual dehazing of urological surgical robots,'' in \textit{Proc. IEEE Int. Conf. Robot. Autom. (ICRA)}, 2025, pp. 4070--4076.

\bibitem{ref26}
Y. Chen, W. Yu, G. Li, et al., ``LL-GaussianImage: Efficient image representation for zero-shot low-light enhancement with 2D Gaussian splatting,'' 2026, arXiv: 2601.15772.

\bibitem{ref27}
Y. Chen, Y. Fang, G. Li, et al., ``LL-GaussianMap: Zero-shot low-light image enhancement via 2D Gaussian splatting guided gain maps,'' 2026, arXiv: 2601.15766.

\bibitem{ref28}
L. Zhu, G. Lin, J. Chen, et al., ``Large images are Gaussians: High-quality large image representation with levels of 2D Gaussian splatting,'' in \textit{Proc. AAAI Conf. Artif. Intell.}, 2025, pp. 10977--10985.

\bibitem{ref29}
Z. Zeng, Y. Wang, T. Guan, et al., ``Instant GaussianImage: A generalizable and self-adaptive image representation via 2D Gaussian splatting,'' in \textit{Proc. IEEE/CVF Int. Conf. Comput. Vis. (ICCV)}, 2025, pp. 27896--27905.

\bibitem{ref30}
X. Zhang, X. Ge, X. Xu, et al., ``GaussianImage: 1000 fps image representation and compression by 2D Gaussian splatting,'' in \textit{Proc. Eur. Conf. Comput. Vis. (ECCV)}, Cham, Switzerland: Springer, 2024, pp. 327--345.

\bibitem{ref31}
C. Jiang, Z. Li, H. Zhao, et al., ``Beyond pixels: Efficient dataset distillation via sparse Gaussian representation,'' 2025, arXiv: 2509.26219.

\bibitem{ref32}
Y. Omri, C. Ding, T. Weissman, and T. Tambe, ``Vision-language alignment from compressed image representations using 2D Gaussian splatting,'' 2025, arXiv: 2509.22615.

\bibitem{ref33}
J. Hu, B. Xia, B. Chen, W. Yang, and L. Zhang, ``GaussianSR: High fidelity 2D Gaussian splatting for arbitrary-scale image super-resolution,'' in \textit{Proc. AAAI Conf. Artif. Intell.}, vol. 39, no. 4, pp. 3554--3562, 2025.

\bibitem{ref34}
B. Kerbl, G. Kopanas, T. Leimk\"{u}hler, and G. Drettakis, ``3D Gaussian splatting for real-time radiance field rendering,'' \textit{ACM Trans. Graph.}, vol. 42, no. 4, p. 139--1, 2023.

\bibitem{ref35}
B. Huang, Z. Yu, A. Chen, A. Geiger, and S. Gao, ``2D Gaussian splatting for geometrically accurate radiance fields,'' in \textit{Proc. ACM SIGGRAPH}, 2024, pp. 1--11.

\bibitem{ref36}
Y. Yan, H. Lin, C. Zhou, et al., ``Street Gaussians: Modeling dynamic urban scenes with Gaussian splatting,'' in \textit{Proc. Eur. Conf. Comput. Vis. (ECCV)}, Cham, Switzerland: Springer, 2024, pp. 156--173.

\bibitem{ref37}
A. Hanson, A. Tu, G. Lin, et al., ``Speedy-splat: Fast 3D Gaussian splatting with sparse pixels and sparse primitives,'' in \textit{Proc. IEEE/CVF Conf. Comput. Vis. Pattern Recognit. (CVPR)}, 2025, pp. 21537--21546.

\bibitem{ref38}
Y. Xu, J. Zhang, Y. Chen, et al., ``PMGS: Reconstruction of projectile motion across large spatiotemporal spans via 3D Gaussian splatting,'' 2025, arXiv: 2508.02660.

\bibitem{ref39}
Y. Xu, J. Zhang, H. Liu, et al., ``PEGS: Physics-event enhanced large spatiotemporal motion reconstruction via 3D Gaussian splatting,'' 2025, arXiv: 2511.17116.

\bibitem{ref40}
X. Zhou, Z. Lin, X. Shan, et al., ``DrivingGaussian: Composite Gaussian splatting for surrounding dynamic autonomous driving scenes,'' in \textit{Proc. IEEE/CVF Conf. Comput. Vis. Pattern Recognit. (CVPR)}, 2024, pp. 21634--21643.

\bibitem{ref41}
K. Cheng, X. Long, K. Yang, et al., ``GaussianPro: 3D Gaussian splatting with progressive propagation,'' in \textit{Proc. Int. Conf. Mach. Learn. (ICML)}, 2024.

\bibitem{ref42}
Y. Liu, C. Luo, L. Fan, et al., ``CityGaussian: Real-time high-quality large-scale scene rendering with Gaussians,'' in \textit{Proc. Eur. Conf. Comput. Vis. (ECCV)}, Cham, Switzerland: Springer, 2024, pp. 265--282.

\bibitem{ref43}
J. Fan, W. Li, Y. Han, T. Dai, and Y. Tang, ``Momentum-GS: Momentum Gaussian self-distillation for high-quality large scene reconstruction,'' in \textit{Proc. IEEE/CVF Int. Conf. Comput. Vis. (ICCV)}, 2025, pp. 25250--25260.

\bibitem{ref44}
T. Yi, J. Fang, J. Wang, et al., ``GaussianDreamer: Fast generation from text to 3D Gaussians by bridging 2D and 3D diffusion models,'' in \textit{Proc. IEEE/CVF Conf. Comput. Vis. Pattern Recognit. (CVPR)}, 2024, pp. 6796--6807.

\bibitem{ref45}
C. Ni, G. Zhao, X. Wang, et al., ``ReconDreamer: Crafting world models for driving scene reconstruction via online restoration,'' in \textit{Proc. IEEE/CVF Conf. Comput. Vis. Pattern Recognit. (CVPR)}, 2025, pp. 1559--1569.

\bibitem{ref46}
C. Ma, J. Zhao, and J. Chen, ``DehazeGS: 3D Gaussian splatting for multi-image haze removal,'' \textit{IEEE Signal Process. Lett.}, 2025.

\bibitem{ref47}
J. Yu, Y. Wang, A. Jiang, et al., ``DehazeGS: Seeing through fog with 3D Gaussian splatting,'' 2025, arXiv: 2501.03659.

\bibitem{ref48}
J. Xu, H. Wang, M. Li, et al., ``DehazeSplat: Unifying unsupervised dehazing and 3D Gaussian splatting for robust reconstruction in hazy scenes,'' SSRN 5601728, 2025.

\bibitem{ref49}
N. Jain, A. Jong, S. Scherer, et al., ``SmokeSeer: 3D Gaussian splatting for smoke removal and scene reconstruction,'' 2025, arXiv: 2509.17329.

\bibitem{ref50}
Q. Zhu, J. Mai, and L. Shao, ``A fast single image haze removal algorithm using color attenuation prior,'' \textit{IEEE Trans. Image Process.}, vol. 24, no. 11, pp. 3522--3533, 2015.

\bibitem{ref51}
B. Li, W. Ren, D. Fu, et al., ``Benchmarking single-image dehazing and beyond,'' \textit{IEEE Trans. Image Process.}, vol. 28, no. 1, pp. 492--505, 2018.

\bibitem{ref52}
K. Chi, Y. Yuan, and Q. Wang, ``Trinity-Net: Gradient-guided Swin transformer-based remote sensing image dehazing and beyond,'' \textit{IEEE Trans. Geosci. Remote Sens.}, vol. 61, pp. 1--14, 2023.

\bibitem{ref53}
C. O. Ancuti, C. Ancuti, and R. Timofte, ``NH-HAZE: An image dehazing benchmark with non-homogeneous hazy and haze-free images,'' in \textit{Proc. IEEE/CVF Conf. Comput. Vis. Pattern Recognit. Workshops (CVPRW)}, 2020, pp. 444--445.

\bibitem{ref54}
Y. Shao, L. Li, W. Ren, et al., ``Domain adaptation for image dehazing,'' in \textit{Proc. IEEE/CVF Conf. Comput. Vis. Pattern Recognit. (CVPR)}, 2020, pp. 2808--2817.

\end{thebibliography}
\end{document}